# The Hand-object Kinematic Model for Bimanual Manipulation

**Jingyi Li*[1]**

Polytechnic University of Catalonia, ETSEIB, SPAIN.
(E-mail: jingyi.li1@upc.edu)

**Highlights**

1. A planar kinematic model for bimanual manipulation is presented. It can compute the fingers configurations based on the desired position of the object.
2. In the hand-object plane, the kinematic solutions from the model can generate valid manipulation strategies in work space.
3. The given object positions are the inputs of our simulations, and the joints values as the outputs can realize bimanual manipulation .

**Abstract**

This paper addresses the planar finger kinematics for seeking optimized manipulation strategies. The first step is to model based on geometric features of linear and rotation motion so that the robot can select the fingers configurations. This kinematic model considers the motion between hands and object. Based on 2-finger manipulation cases, this model can output the strategies for bimanual manipulation. For executing strategies, the second step is to seek the appropriate values of finger joints according to the ending orientation of fingers. The simulation shows that the computed solutions can complete the relative rotation and linear motion of unknown objects.

## 1. Introduction

The significance of manipulation strategy shows in properly approaching the desired position of grasped objects. During the approaching process, the complexity of computing inverse solutions for multi-finger manipulation can be found. The inverse kinematics (IK) affects the accuracy of path planning and error tracking for the manipulation in robotic hands [1]. There may be some cases of missing solutions, multiple solutions or have not solutions [2] when researching inverse kinematics. These cases meant the fingertips can not touch the object surface for the in-hand manipulation. Usually, for the multi-finger manipulation framework, researchers will compute the inverse solutions through the fingers Jacobian matrix [3]. It simplifies the process of modelling for the speed or force control of the manipulation. However, the inverse computation can be complex and coupled from an object velocity to each finger velocity. The reference [4][5] emphasize to decompose the manipulation framework. It means researchers can only consider the inverse computation based on the contact point on the fingertip which can always grip the object. Also, the decomposition means the computation can be separated into a planar map involves object pose and fingertip pose. Here the constraints are important to seek the optimal combination of two inverse solutions.

Usually, the inverse solution which is computed by the Monte Carlo method is non-linear and not unique. Here researchers need to define the random density function or motion cost function for seeking the optimal solution [6]. The random samples can be the joints values after analyzing and selecting[7]. Besides, the clustering or dimension reduction can simplify these random samples from 3D to lower dimension [8]. For the kinematic simulation, the algorithm design for approaching or seeking is necessary [9]. During the manipulation, robotic hands will create a large number of discrete position samples. For the proper kinematic solution, there are 3 methods. First, the algebraic approach for the finger to follow the movement of contact point so that the object can be well grasped. Then, it is the iterative approach for avoiding the singularity of finger in the work space. Third, it is the geometric approach which can be suitable for the fuzzy logic or neural networks [10].

This paper describes the related works of kinematic modelling and in-hand manipulation for the dual-hand robots in Sec. 2, and the problem statement of robotic fingers computation for bimanual manipulation in Sec. 3. Besides, this paper will describe the approach of manipulation kinematics modelling in Sec. 4. After that, Matlab based simulations will be discussed in Sec 5, and the conclusion with the future works will be discussed in Sec. 6.

## 2. Related works
### 2.1. Finger-object model for computing solution.

The finger-object model can estimate the states of the robotic dynamics or the kinematics of a finger-object system [11]. Researchers [12] proposed that planar finger-object model can solve most of manipulation tasks. They [13] defined virtual working plane or virtual contact points based on kinematics. In their methods [14], the robot can estimate and guarantee the rolling manipulation by tracking the contact points of the finger without object information. The weakness of the planar finger-object model is that the manipulation stability and available strategies are limited in these methods.

Approaching a position from multiple fingers meant a large number of joint values. Here the Monte Carlo method can provide the values samples in numerical simulation [15]. We can use some of these values to configure the the bimanual manipulation.

Thus, these samples can be the foundation of optimizing the solutions and can extend the manipulation strategies.

Based on some of proper joints values in the finger-object model, it is possible to formalize and use learning techniques [16]. The first is to learn based on the object positions in the working space via the PCA (principal component analysis). The second is to learn based on the iterated valid fingers configurations. The key of the learning is how to estimate the relation between fingers and object movement. When the robot learned the features from the object path or the finger working plane, the robot can be heuristic to seek the analytical solutions and then to be object-centered.

**2.2. In-hand Manipulation.**

In the last decades [17], researchers believed that the valid grasp gestures and dexterity are the foundation of in-hand manipulation. And they tried to transfer the grasp gestures into the manipulation configuration based on the taxonomy of grasp gestures. The valid grasp gestures for the manipulation in the inverse kinematics should be solvable. For the most of manipulation situation, the initial grasp gestures are known. Also, according to the manipulation tasks, the desired object positions can be known via the path planning. However, there are few of gestures for dual-hand manipulation. Also, the inverse kinematic computation is more challengeable. Before inputting the inverse solution into the robotic control or dynamic model, we hope to seek the optimized fingertip positions. It should be effective, efficient and appropriate for manipulation. Under these positions, the robot can constantly receive the incremental parameters for approaching the relative goal pose. Here the success rates and the accuracy of manipulation is important [18]. In our works, the manipulation strategies are presented based on 3 basic assumptions:
   1. The unknown object is rigid or finite deformable during the manipulation.
   2. The unknown object is initially well grasped by the two robotic hands.
   3. The wrists of the two hands is fixed or minimally moved during the motion sequence of fingers.

**3. Problem statement.**

For enabling the robot to move the object by 2 robotic hands, we need to compute inverse analytical solution of each finger joint. Here, optimized and appropriate joints values of fingers will be selected according to the relative motion of the object and the contact points. For each contact point, its position should be linearly correlated with the movement of the object position. Here, the world absolute reference frame is established on the robotic palm. Then, each finger will obtain the joints poses and establish the planar relative reference frames. We can initially compute the positions of contact point according to the relative motion of the objects. Then, the contact point determines the planar object poses. Besides, based on the palm, the fingers reference frames and the object reference frame are derived for expressing different motions. Before the manipulation, the two hands have well grasped the object. Thus, the initial configurations for manipulation are known by the robot. In this paper, 2-finger cases, as the represent of bimanual manipulation, will be considered.

## 4. Kinematic modelling
### 4.1 Linear and angular object movement.

Figure 1 (a) shows possible linear movement of unknown object which depends on the relative motion in the flexion-extension work plane. During the manipulation, from the initial pose $P_o`$ to the desired object pose $P_o^{lin}$, exists:

$$P_o^{lin} = P_o` + \Delta P_o^{lin} = \begin{bmatrix} x_o` \\ y_o` \\ \varphi_o` \end{bmatrix} + \begin{bmatrix} \Delta x_o \\ \Delta y_o \\ \Delta \varphi_o \end{bmatrix} = \begin{bmatrix} x_o \\ y_o \\ \varphi_o \end{bmatrix} \quad (1)$$

Where $\Delta x_o$ and $\Delta y_o$ are the movements values from flexion-extension work plane. For executing this relative motion by 2 fingers, the desired positions of the contact points, $c_1$ and $c_2$, can be:

$$\begin{cases} c_1 = [x_1` + \Delta x_o, \; y_1` + \Delta y_o]^T \\ c_2 = [x_2` + \Delta x_o, \; y_2` + \Delta x_o]^T \end{cases}$$

$$\begin{cases} c_1` = [x_1`, y_1`]^T \\ c_2` = [x_2`, y_2`]^T \end{cases} \quad (2)$$

Where $c_1`$ and $c_2`$ are the initial positions of the contact points. Besides, in Fig. 1 (b), the object is rotated in the object reference frame. For executing this angular movement base on the robotic palm, the desired positions of the contact points can be denoted as:

$$\begin{cases} c_1 = [x_o`, y_o`]^T + R_o(\beta) \cdot [x_1` - x_o`, \; y_1` - y_o`]^T \\ c_2 = [x_o`, y_o`]^T + R_o(\beta) \cdot [x_2` - x_o`, \; y_2` - y_o`]^T \end{cases}$$

$$R_o(\alpha) = \begin{bmatrix} \cos\beta & -\sin\beta \\ \sin\beta & \cos\beta \end{bmatrix} \quad (3)$$

Where $R_o(\beta)$ is the rotation matrix and $\beta$ is the desired rotation angle in the relative reference frame of the object.

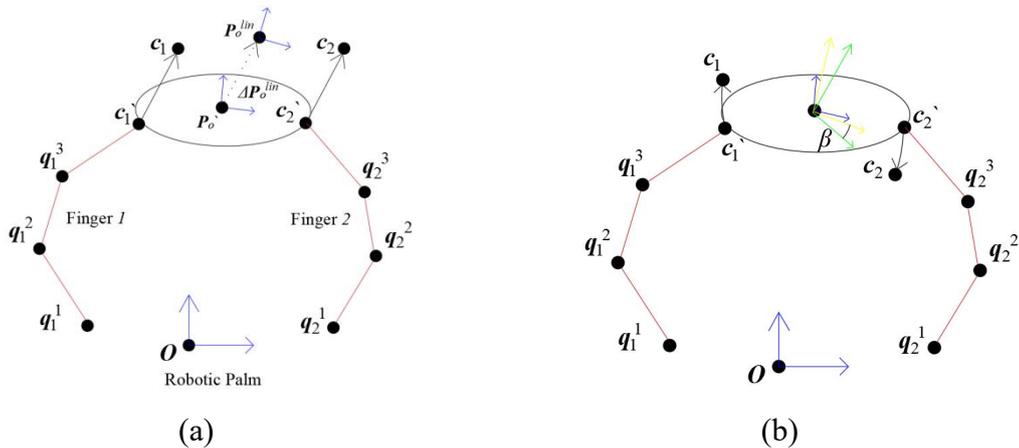

(a)  (b)

Fig. 1 Manipulation modes for 2-finger case

*Where $q_1^3$ means the configuration of joint 3 in finger 1. **O** is the coordinate origin of flexion-extension work plane for expressing the object movement.*

## 2 Steps for 2-finger IK solutions.

For executing the object motions, there are 6 steps for seeking inverse solutions from object to each joint, exists:

**Step 1:** In flexion-extension work plane, obtaining initial configurations of the 2 fingers as the Figure 2 shown.

**Step 2:** Compute the desired positions of the contact points according to Equation (2) or Equation (3).

**Step 3:** Compute the norms, $s_1$ and $s_2$, between the contact points and the base joint $q_1^1$ and $q_2^1$ in Figure 2 as:

$$\begin{cases} s_1 = \sqrt{(x_1 - x_1^1)^2 + (y_1 - y_1^1)^2} \\ s_2 = \sqrt{(x_2 - x_2^1)^2 + (y_2 - y_2^1)^2} \end{cases} \quad (4)$$

In the flexion-extension work plane, $x_1^1$ and $y_1^1$, $x_2^1$ and $y_2^1$ are the positions values of $q_1^1$ and $q_2^1$. Also, for the position values $x_1$ and $y_1$, and $x_2$ and $y_2$, exists:

$$\begin{cases} c_1 = [x_1, y_1]^T \\ c_2 = [x_2, y_2]^T \end{cases} \quad (5)$$

**Step 4:** Compute the angle $\alpha_1$ and $\alpha_2$ in each relative reference frame of the finger as Figure 2 shown.

$$\begin{cases} \arccos \alpha_1 = \frac{x_1 - x_1^1}{s_1} \\ \arccos \alpha_2 = \frac{x_2 - x_2^1}{s_2} \end{cases} \quad (6)$$

**Step 5:** Compute the joints values according to the analytical solution. Assume the fingers joints are sufficient and not redundant, then, the ending orientation can be fixed during the manipulation. Thus, the relation between desired ending orientations, $\varphi_{c1}$ and $\varphi_{c2}$, and the joints values can be:

$$\begin{cases} \varphi_{c1} = \varphi_{c1}` = \theta_1^1 - \theta_1^2 - \theta_1^3 \\ \varphi_{c2} = \varphi_{c2}` = \theta_2^1 + \theta_2^2 + \theta_2^3 \end{cases} \quad (7)$$

Furthermore, in Figure 2, we can see $\alpha_1 < \alpha_2$. Then, for the joints values of $q_1^1$ and $q_2^1$.

$$\begin{cases} \theta_1^1 = (\frac{1}{3}\varphi_{c1} + \alpha_1) \\ \theta_2^1 = (\alpha_2 - \frac{1}{3}\varphi_{c2}) \end{cases} \quad (8)$$

**Step 6:** Let $\theta_1^2 = \theta_1^3 = \frac{1}{3}\varphi_{c1}$ and $\theta_2^2 = \theta_2^3 = \frac{1}{3}\varphi_{c2}$, then, the Equation (9) is obtained.

$$\begin{cases} s_1 \cdot \cos \alpha_1 = l_1^1 \cdot \cos(\alpha_1 + \frac{1}{3}\varphi_{c1}) + l_1^2 \cdot \cos(\alpha_1 + \frac{2}{3}\varphi_{c1}) + l_1^3 \cdot \cos(\alpha_1 + \varphi_{c1}) \\ s_1 \cdot \sin \alpha_1 = l_1^1 \cdot \sin(\alpha_1 + \frac{1}{3}\varphi_{c1}) + l_1^2 \cdot \sin(\alpha_1 + \frac{2}{3}\varphi_{c1}) + l_1^3 \cdot \sin(\alpha_1 + \varphi_{c1}) \\ s_2 \cdot \cos \alpha_2 = l_2^1 \cdot \cos(\alpha_2 - \frac{1}{3}\varphi_{c2}) + l_2^2 \cdot \cos(\alpha_2) + l_2^3 \cdot \cos(\alpha_2 + \frac{1}{3}\varphi_{c2}) \\ s_2 \cdot \sin \alpha_2 = l_2^1 \cdot \sin(\alpha_2 - \frac{1}{3}\varphi_{c2}) + l_2^2 \cdot \sin(\alpha_2) + l_2^3 \cdot \sin(\alpha_2 + \frac{1}{3}\varphi_{c2}) \end{cases} \quad (9)$$

Based on the Equation (9), the algorithm for 2-finger IK can be:

---
*Algorithm 1:* IK solutions of 2-finger joints
---

*Input*:

initial object pose $P_o` = [x_o`, y_o`, \varphi_o`]^T$; desired object pose $P_o = [x_o, y_o, \varphi_o]^T$;

initial position of contact points $c_i`$, where $i \ni [1,2]$; $j \ni [1,2,3]$;

link length $l_i^j$; ending orientation $\varphi_{ci}$;

*Output*:

Joint values $\theta_1^1, \theta_1^2, \theta_1^3, \theta_2^1, \theta_2^2, \theta_2^3$

1: *Compute $s_1$ and $s_2$ according to Eq. (4)*;

2: *Compute $\alpha_1$ and $\alpha_2$ according to Eq. (5)*;

3: **If** $\varphi_o` = \varphi_o$ **then**

4: | *Compute $c_i$ according to Eq. (2)*;

5: | **else** *Compute $c_i$ according to Eq. (3)*;

6: **If** $\alpha_1 < \alpha_2$ **then**

7: | $\theta_1^1 \leftarrow (\frac{1}{3}\varphi_{c1} + \alpha_1)$; $\theta_2^1 \leftarrow (\alpha_2 - \frac{1}{3}\varphi_{c2})$;

8: | $k_1 \leftarrow \frac{1}{3}\varphi_{c1}$; $k_2 \leftarrow \theta_2^1$;

9: | **else** $\theta_1^1 \leftarrow (\alpha_1 - \frac{1}{3}\varphi_{c1})$; $\theta_2^1 \leftarrow (\frac{1}{3}\varphi_{c2} + \alpha_{c2})$;

10: | $k_1 \leftarrow \theta_1^1$; $k_2 \leftarrow \frac{1}{3}\varphi_{c2}$;

11: $\theta_1^2 \leftarrow k_1$; $\theta_1^3 \leftarrow k_1$; $\theta_2^2 \leftarrow k_2$; $\theta_2^3 \leftarrow k_2$;

12: $\varphi_{c1} \leftarrow (\theta_1^1 - \theta_1^2 - \theta_1^3)$; $\varphi_{c2} \leftarrow (\theta_2^1 + \theta_2^2 + \theta_2^3)$;

13: **If** $s_i \cdot \cos \alpha_i < l_i^1 \cdot \cos(\theta_i^1) + l_i^2 \cdot \cos(\theta_i^1 + \theta_i^2) + l_i^3 \cdot \cos(\theta_i^1 + \theta_i^2 + \theta_i^3)$

14: | **then** $\varphi_{ci} \leftarrow \varphi_{ci} + \frac{1}{3}\varphi_{ci}$;

15: | *Reposition for palm pose $O$*;

   # *Adjust ending orientation for approaching*

16: | **else** $\varphi_{ci} \leftarrow \varphi_{ci} - \frac{1}{3}\varphi_{ci}$;

17: | *Reposition for palm pose $O$*;

18: **Return** $c_i$; **end**.

---

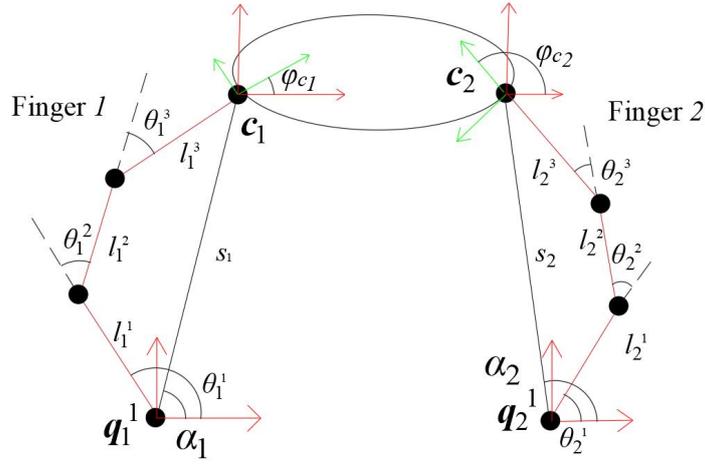

**Fig. 2 Planar 2-finger manipulation case**

Where $\theta_1^3$ means the value of joint 3 in Finger 1 and $l_1^3$ means the length of link 3 in Finger 1. The red arrows meant the relative reference frames of the 2 fingers.

## 5 Bimanual simulations.

### 5.1 2-finger cases.

For our planar kinematic simulation based on Fig. 2, the initial parameters are set as:

**Table 1** Inputs for the simulation

| Contact points positions (mm) | | | Coordinates of the base joints (mm) | | |
|---|---|---|---|---|---|
| $c_1$ | $x_1$ | -5 | $q_1^1$ | $x_1^1$ | -15 |
| | $y_1$ | 70 | | $y_1^1$ | 0 |
| $c_2$ | $x_2$ | 30 | $q_2^1$ | $x_2^1$ | 15 |
| | $y_2$ | 75 | | $y_2^1$ | 0 |

Then, according to Equation (4.2.1) and (4.2.3), exists:

**Table 2** Computed parameters in the simulation

| Finger 1 | | Finger 2 | |
|---|---|---|---|
| $s_1$ (mm) | 70.7 | $s_2$ (mm) | 76.5 |
| $\alpha_1$ (°) | 81.9 | $\alpha_2$ (°) | 78.7 |
| $\varphi_1$ (°) | 30 | $\varphi_2$ (°) | 120 |

Based on Table 2 and the Algorithm 1, we can obtain computed results. Besides, based on the Matlab/robotic toolbox, we can obtain the simulation results as:

Table 3 Simulation results

| Desired positions (mm) | | Initial joints values positions (°) | | | | Computed joints values positions (°) | | | |
|---|---|---|---|---|---|---|---|---|---|
| $c_1$( -5, 70) | $c_2$( 30, 75) | Finger 1 | | Finger 2 | | Finger 1 | | Finger 2 | |
| Computed positions | | $q_1$ | 180 | $q_1$ | 0 | $q_1$ | 180 | $q_1$ | 0 |
| $c_1$(-5.04, 69.9) | $c_2$(29.9, 75.1) | $q_3$ | 78 | $q_3$ | 47 | $q_3$ | 50 | $q_3$ | 40 |
| Actual positions | | $q_5$ | 45 | $q_5$ | 45 | $q_5$ | 50 | $q_5$ | 40 |
| $c_1$(-3.09, 67.5) | $c_2$(28.2, 74.5) | $q_6$ | 37 | $q_6$ | 28 | $q_6$ | 50 | $q_6$ | 40 |

Then, the relation between actual positions of contact points and the joints values can be plotted in Figure 3 as:

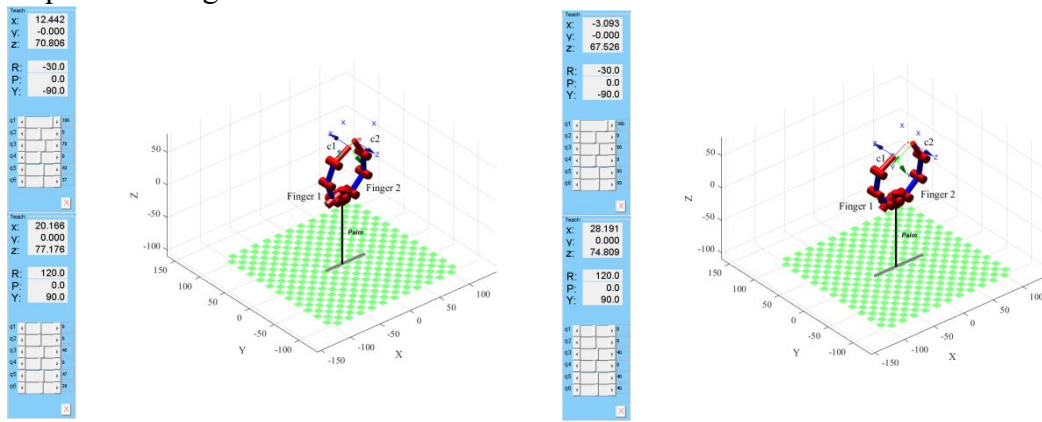

(a) Start  (b) End

**Fig. 3 2-finger manipulation results**

In Table 3 and Figure 3, the $q_1$ of Finger 1 is 180 due to the symmetry with the Finger 2 in palm absolute reference frame. Both of $q_2$ and $q_4$ are the joints which can not rotate in our working plane and they are set as null. The distance between the world coordinate origin and each base joint is 15m. After joint values are inputted in the Matlab/robotic toolbox, in Figure 3, we can see the RPY parameters of the ending effector (defined as contact point). Here, the values of P and Y axis can be neglected due to the planar kinematics and R represents the ending orientation.

**5.2 Errors analysis**

Based on 2-finger manipulation cases, the errors between desired, computed and actual positions of contact points can be analyzed. When 2 fingers are well grasping a sphere object whose radius is 15mm, there are 2 strategies that can be determined.

   a: the simultaneous strategy: increase or decrease the joint values between 2 fingers so that the object can move linearly and vertically as Fig. 4 (a) and 4 (b) shown.
   b: the interleaved strategy: once the joint values of a finger are increased, the other finger should be interleaved to decrease its joint values for the rotation or translational motion of the object in the working plane.

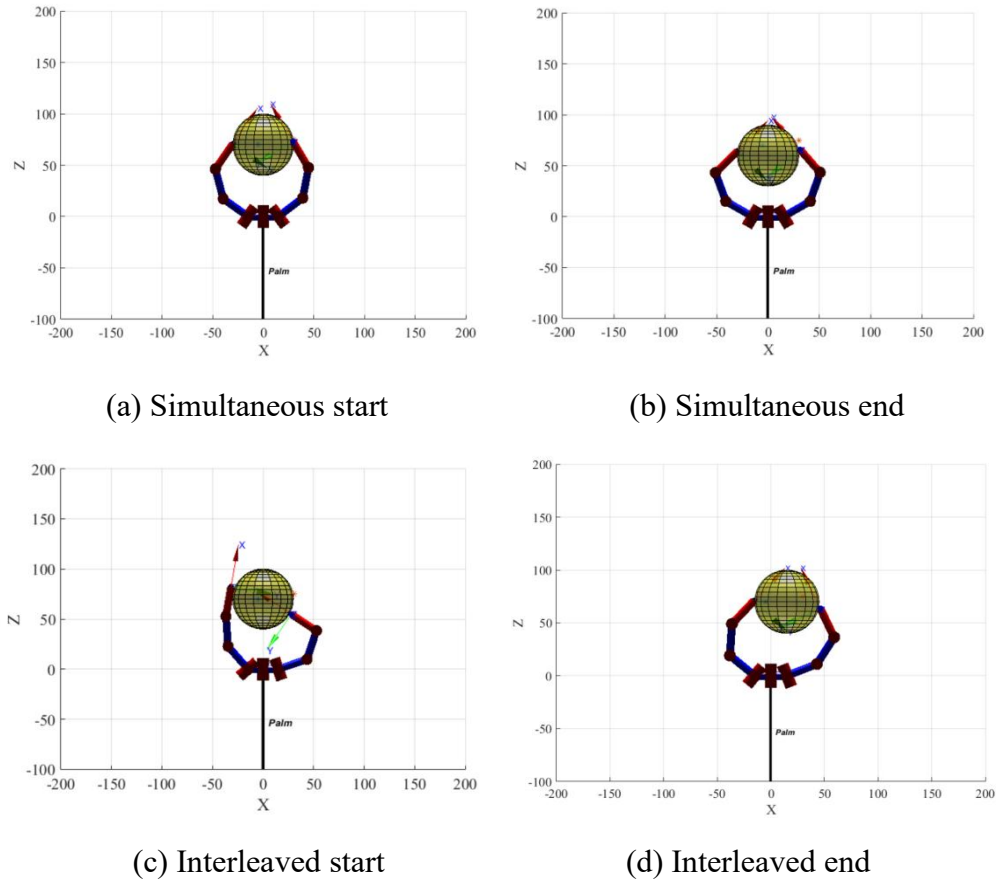

(a) Simultaneous start  (b) Simultaneous end

(c) Interleaved start  (d) Interleaved end

Fig. 4 The strategies for the 2 fingers

Fig. 4 (a) and 4 (b) shown a 10 mm vertical motion of the object for 15 s. Between the desired and actual positions of the contact points, the errors in 7 s are the top, with the Finger 1 is 2.05mm and the Finger 2 is 1.94 mm. Figure 4 (c) shows 15-degree angular movement of the object for 15s. The errors in 11s are the top on the *z* coordinate with the Finger 1 is 2.66mm and the Finger 2 is 2.17 mm. Figure 4 (d), shows 10mm translational motion for 15s. The errors in 15s are the top on the *x* coordinate with the Finger 1 is 1.98 mm and the Finger 2 is 2.10 mm. For these simulations, our computed errors are all near the desired positions and their errors are less than 0.71 mm, the desired and actual errors are less than 2.66 mm.

## 6. Conclusion and future direction

In this paper, we presented a method for using robotic fingers to move a grasped object. This method is based on planar kinematics and can be applied to the bimanual manipulation. This method can determine the manipulation motion based on once teach of the grasp configuration. After desired pose of the object is set, the fingers configurations for interleaved or simultaneous manipulation strategy can be found. Two strategies promoted the dexterous and robust manipulation. We simulated 20 initial configurations of each finger for the 2 strategies in 15s and then the fingers completed the manipulation for sphere object to approach the desired poses.

Future works need the tests on more unknown objects with complex surface. These objects can provide more manipulation cases for verifying the model accuracy. Besides, this analytical kinematic question needs to be verified in the control model for tracking the robustness when unknown objects moving.

**References.**

[1] J. Wang, C. Gosselin and L. Cheng. Modeling and simulation of robotic systems with closed kinematic chains using the virtual spring approach. Multibody System Dynamics, Vol. 7, No.2, pp. 145-170, 2002.

[2] M. Liarokapis and M. Dollar. Combining analytical modeling and learning to simplify dexterous manipulation with adaptive robot hands. IEEE Transactions on Automation Science and Engineering, Vol. 11, No. 10, pp. 1361-1372, 2018.

[3] I. Ozawa, S. Arimoto, S. Nakamura et. al. Control of an object with parallel surfaces by a pair of finger robots without object sensing. IEEE Transactions on Robotics, Vol. 21, No. 5, pp. 965-976, 2005.

[4] T. Fukuda, K. Mase and Y. Hasegawa. Robot hand manipulation by evolutionary programming. In Proc. IEEE Int. Conf. Robotics and Automation (ICRA), vol.3, pp. 2458-2463, 1999.

[5] D. Aaron, S. Coogan, M. Egerstedt, et al. Control barrier functions: Theory and applications. IEEE 18th European Control Conference (ECC), pp. 1-19, 2019.

[6] L. Han and J. C. Trinkle. Dextrous manipulation by rolling and finger gaiting. In IEEE International Conference on Robotics and Automation, pp. 730–735, 1998.

[7] M. Liarokapis, V. Minas and A. M. Dollar . Learning Task-Specific Models for Dexterous, In-Hand Manipulation with Simple, Adaptive Robot Hands." IEEE/RSJ International Conference on Intelligent Robots and Systems (IROS) IEEE, 2016.

[8] S. Qiu and M. Kermani. A new approach for grasp quality calculation using continuous boundary formulation of grasp wrench space. Mechanism and Machine Theory, 168, pp. 104-119, 2022.

[9] H. Amor, O. Kroemer, U. Hillenbrand , et al. Generalization of human grasping for multi-fingered robot hands. 2012 IEEE/RSJ International Conference on Intelligent Robots and Systems. IEEE, pp. 2043-2050, 2012.

[10] L. Harillo, Antonio P. González, J. Starke, et al. The Anthropomorphic hand assessment protocol (AHAP). Robotics and Autonomous Systems, Vol. 121, pp.103-128, 2019.

[11] H. Park and C Lee, Dual-arm coordinated-motion task specification and performance evaluation, in IEEE/RSJ International Conference on Intelligent Robots and Systems (IROS). IEEE, pp. 929–936, 2016.

[12] R. Ozawa, S. Arimoto, S. Nakamura, et. al. Control of an object with parallel surfaces by a pair of finger robots without object sensing. IEEE Transactions on Robotics, Vol. 21, No. 5, pp.965-976.

[13] M. Raymond, and M. Dollar. An underactuated hand for efficient finger-gaiting-based dexterous manipulation. IEEE International Conference on Robotics and Biomimetics (ROBIO 2014). IEEE, pp. 10-16, 2014.

[14] A. Montaño and R. Suárez. Dexterous manipulation of unknown objects using virtual contact points. Robotics, Vol. 8, No. 4, pp.86-196, 2019.

[15] S. Yahya, M. Moghavvemi and H. Mohamed. Geometrical approach of planar hyper-redundant manipulators: Inverse kinematics, path planning and workspace. Simulation Modelling Practice and Theory, Vol. 19, No. 1, pp.406-422, 2011.

[16] I. Bullock, M. Raymond and M. Dollar. A hand-centric classification of human and robot dexterous manipulation. IEEE Transactions on Haptics, Vol. 6, No. 2, pp.129-144. 2012.

[17] L. Basañez and R. Suárez. Teleoperation. In Springer Handbook of Automation, pp. 449–468. Springer-Verlag, 2009.

[18] N. García, Motion planning using synergies: application to anthropomorphic dual-arm robots. Ph.D. Thesis of UPC, 2019.